\documentclass[review]{elsarticle}
\usepackage{lineno,hyperref}
\usepackage{algorithm}
\usepackage{algpseudocode}
\usepackage{graphics}
\usepackage{epsfig}
\usepackage{amssymb}
\usepackage{color}
\usepackage{amsmath}
\usepackage{graphicx}
\usepackage{epstopdf}
\usepackage{ulem}
\usepackage{setspace}
\usepackage{amssymb}
\usepackage{lipsum}
\usepackage{subcaption,graphicx}
\usepackage{times}
\usepackage{ulem}
\usepackage{mathrsfs}
\usepackage[table]{xcolor}
\usepackage{booktabs}
\usepackage{makecell}
\usepackage{colortbl}
\usepackage{multirow}
\usepackage{array}
\usepackage{balance}
\definecolor{mygray1}{gray}{.9}
\definecolor{mygray2}{gray}{.5}
\modulolinenumbers[5]
\journal{Journal of \LaTeX\ Templates}

{}
{}
{}
\begin{document}

\begin{frontmatter}

\title{Enhance the Motion Cues for Face Anti-Spoofing using CNN-LSTM Architecture}
\author{Xiaoguang Tu\fnref{footnote1}}
\author{{\color{black}Hengsheng Zhang}\fnref{footnote1}}
\author{Mei Xie\fnref{footnote1}}
\author{Yao Luo\fnref{footnote1}}
\author{{\color{black}Yuefei Zhang}\fnref{footnote2}}
\author{Zheng Ma\fnref{footnote1} \corref{correspondingauthor}}
\address{No.2006, Xiyuan Ave, West Hi-Tech Zone, 611731 Chengdu, Sichuan, P.R.China}
\fntext[footnote1]{School of Communication and Information Engineering, University of Electronic Science and Technology of China.}
\fntext[footnote2] {Chongqing Institute of Public Security Science and Technology.}
\cortext[correspondingauthor]{Corresponding author, Email: zma\uline{ }uestc@outlook.com}

\begin{abstract}
Spatio-temporal information is very important to capture the discriminative cues between genuine and fake faces from video sequences. To explore such a temporal feature, the fine-grained motions (e.g., eye blinking, mouth movements and head swing) across video frames are very critical. In this paper, we propose a joint CNN-LSTM network for face anti-spoofing, focusing on the motion cues across video frames. We first extract the high discriminative features of video frames using the conventional Convolutional Neural Network (CNN). Then we leverage Long Short-Term Memory (LSTM) with the extracted features as inputs to capture the temporal dynamics in videos. To ensure the fine-grained motions more easily to be perceived in the training process, the eulerian motion magnification is used as the preprocessing to enhance the facial expressions exhibited by individuals, and the attention mechanism is embedded in LSTM to ensure the model learn to focus selectively on the dynamic frames across the video clips. Experiments on Replay Attack and MSU-MFSD databases show that the proposed method yields state-of-the-art performance with better generalization ability compared with several other popular algorithms.
\end{abstract}


\end{frontmatter}


\section{Introduction}
As a convenient biometrics based authentication, automatic face recognition has attracted increasing attention during the past decades owing to it's convenience and high-efficiency on access control applications. Despite significant recent advances, the security of face recognition systems is still challenged. Face recognition systems are vulnerable to spoofing attacks with printed photos or relayed videos if protective measures are not implemented. Generally, a practical face recognition system demands not only high recognition performance, but also the capability of distinguishing the attackers (fake faces) from real persons (genuine faces).

In biometric based face recognition systems, spoofing attacks are usually perpetrated using photographs, replayed videos and forged mask. Fig. 1 shows some genuine and fake face images from MSU-MFSD \cite{Wen2015Face}, where we can see no obvious visual cues are available to pick the fake images from the gallery. However, considering that many of the spoofing attacks are carried out by replaying a recording video, the motion patterns contained in the genuine and the spoofing video sequences may be different. For example, facial motions such as eye blinking and mouth movement are not contained in video frames of photograph, while exist in video frames of real person and replayed attack videos. Meanwhile, additional motion patterns such as hand-trembling is inevitable brought in the replayed attack videos, if the attack equipment is handheld. These motion cues are very valuable to distinguish valid access of attempted attacks, and also the relative motion between the face region and the background can be helpful. Generally, the methods using motion cues are expected to have better generalization ability than other methods, considering the motion patterns are not vary due to replay attacks.
\begin{figure}[h]
    \hspace{0.61em}
    \includegraphics[width = 11.6cm, height=4.8cm]{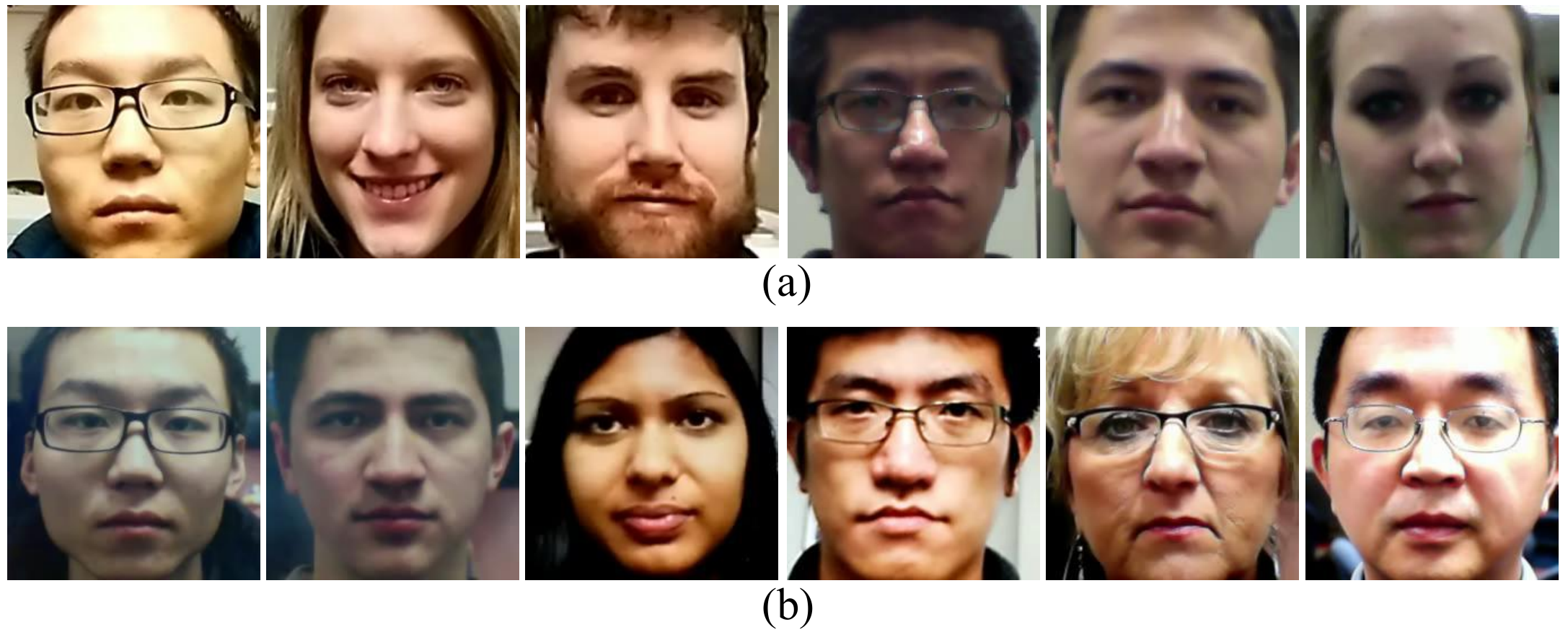}%

\begin{spacing}{0.9}
{\footnotesize{Fig.1: Genuine and fake face images selected from MSU-MFSD database. (a) Genuine images. (b) Fake images.}}
\end{spacing}
\end{figure}

In this paper, we proposed a joint CNN-LSTM architecture for the purpose of robust face anti-spoofing, focusing on the motion cues across video frames. We have also proposed several strategies to ensure the motion cues be used sufficiently. Two publicly databases, the Replay Attack \cite{Chingovska2012On} and MSU-MFSD have been used to evaluate the performance of the proposed method, using the official protocols. Cross-database experiments are also performed to evaluate the generalization ability. The main contributions of our work can be summarized as follows:
\begin{itemize}
\item We propose to use the LSTM-CNN architecture to learn the temporal features and make use of the motion cues across video frames for face anti-spoofing.
\item The Eulerian motion magnification approach is used as the preprocessing to enhance the facial expressions exhibited by individuals.
\item The attention mechanism is embedded in the LSTM to selected dynamic key video frames.
\item A confusion loss layer based on LSTM loss and CNN loss is created to balance the learning level of CNN and LSTM for the purpose of better generalization ability.
\end{itemize}

The remainder of this paper is organized as follows. Previous related works on face anti-spoofing are discussed in Sect. 2. We describes the proposed framework in Sect. 3. The discussion and performance evaluation are presented in Sect. 4. Finally, we draw the conclusion in Sect. 5.

\section{Related Works}
\label{sec:1}
Recently, a large number of approaches have been proposed in the literature to detect spoofing attacks based on photographs, replayed videos and forged masks \cite{Wen2015Face,Yang2014Learn,Alotaibi2017Deep,Lucena2017Transfer,Graves2014Generating,Graves2014Hybrid}. Depending on the cues be used, existing face anti-spoofing methods could be roughly categorized into two groups: static approaches and dynamic approaches.

Static approaches are mainly based on the analysis on texture differences between live and spoof face image from the perspective of surface reflection and material differences. These methods perform spoof detection using a single face image, and thus have relatively fast response. In \cite{Li2015Live}, the researches utilized the difference of structural texture between 2D images and 3D images to detect spoofing attacks based on the analysis of Fourier spectra, where the reflections of light on 2D and 3D surfaces result in different frequency distributions. Tan et al. \cite{Tan2010Face} used a variational retinex-based method and the DoG filters to extract the reflectance features on face images to distinguish fake images from geunine images. In \cite{Maatta2011Face}, Maatta et al. extracted the texture of 2D images using the multi-scale local binary pattern (LBP) to generate a concatenated histogram which was fed into a SVM classifier for the classification of fake and genuine faces. They showed that concatenation of three Local Binary Patterns (LBP) descriptors of different configurations is more efficient than local phase quantization as well as Gabor wavelet based descriptor for print attack spoofing detection. In their later work \cite{Maatta2012Face}, a score level fusion approach was proposed using LBP, histogram of oriented gradients and Gabor wavelets computed from the local blocks of face images. The authors reports $0\%$ Half Total Error Rate (HTER) on the Print Attack dataset. To include the temporal features which described the faces' dynamic structure for face anti-spoofing, Pereira et al. \cite{Pereira2012LBP} proposed the LOP-TOP, considering three orthogonal planes intersecting the center of a pixel in the $XY$ direction, $XT$ direction and $YT$ direction,where $T$ is the time axis. According to their experimental results, multi-resolution LBP-TOP with SVM classifier achieved the best HTER OF $7.6\%$ on the Replay Attack dataset. In the work \cite{Boulkenafet2015Face}, Boulkenafet et al. analyze the joint color-texture information from the luminance and the chrominance channels using a color local binary pattern descriptor and showed excellent results according to their experimental results.

Dynamic approaches make use of the liveness information across the input video frames, such as eye blinking, lip movement and head movement \cite{Kollreider2007Real}. Some dynamic method include "intrusive interactions", in which the user is forced to follow some instructions. Pan et al. \cite{Pan2007Eyeblink} detected the eye blinking for anti-spoofing using a non-intrusive method, a conditional random field was constructed to model different stages of eye blinking. In \cite{Kollreider2008Verifying}, Kollreider et al. presented an approach to detect spoofing attacks by combining eye blinking and mouth movement with 3D properties of face images. Facial motions such as eye blinking and head movement were used to determine liveness by some participants of the IJCB facial spoofing competition \cite{Chakka2011Competition}. In \cite{Shreve2011Macro}, Shreve et al. proposed a temporal stain metric computed from optical flow patterns on the facial regions to recognize the subtle facial features in video sequences for the use of detecting spoofing attacks. In the later work \cite{bharadwaj2014face}, Bharadwaj proposed to enhance the micro- and macro facial expressions for more robust detection of spoofing attacks. Two feature extraction algorithms, a configuration of local binary pattern and motion estimation using histogram of oriented optical flow, were used to encode texture and motion (liveness) properties respectively. Their work yields state-of-the-art performance and robust generalizability with low computational complexity according their experimental results.

Apart from these hand-crafted features, some approaches based on CNNs \cite{Yang2014Learn,Alotaibi2017Deep,Lucena2017Transfer} have also been proposed to automatically learn features for face anti-spoofing since the hand-crafted features are designed specifically and a new dataset may result in bad performance. CNNs have been successfully applied to various problems such as image classification \cite{Krizhevsky2012ImageNet}, object recognition\cite{Ren2017Faster} and semantic segmentation \cite{Long2015Fully}. The liveness face detection tasks based on video clips can be regarded as the problems of videos classification instead of images classification with a temporal dimension added. Xu et al. firstly introduced the LSTMs for the use of face anti-spoofing in \cite{Xu2016Learning}. LSTMs have been use to achieve state-of-art performance in several tasks such as sequence generation \cite{Graves2014Generating}, speech recognition \cite{Graves2014Hybrid} and video description \cite{Yao2015Describing} and have shown great power in learning patterns along time series. They proposed to have CNNs underlying the LSTMs, where the local and dense property from convolution operation could be leveraged and the temporal feature across frames can be learned and stored in LSTM units. A limitation to their work is that the spatial locations of receptive fields are mixed in fully connected layers, and neuron activations from fully connected layers do not encode spatial information \cite{Zou2014Generic,Branson2014Bird,Liu2015The}, therefore the features from fully connected layers indicates no location information which is valuable for capturing motion cues.

\section{The proposed framework}
Our work focusing on the motion cues across video frames. A major difference between convolutional and fully connected layer activations is that the former is embedded with rich spatial information while the latter is not \cite{Liu2015The}. To ensure the motions cues adequately be used, we use the features from convolutional layers instead of the full connected layers. Eulerian motion magnification and attention mechanism \cite{Mnih2014Recurrent,Bahdanau2014Neural} are used to enhance the motion cues. A confusion loss layer based on LSTM loss and CNN loss is created to balance the learning level of CNN and LSTM.

\subsection{Magnification of Facial Expressions}
Eulerian motion magnification has been proposed in \cite{Wu2012Eulerian} to reveal temporal variations in videos that are difficult or impossible to see with the naked eye and display them in an indicative manner. This Eulerian based method could successfully reveals informative signals and amplifies small motions in real-world videos.

The motion magnification methods takes a standard sequence as input, and applies spatial decomposition, followed by temporal filtering to the frames. Let $I(x,y,t)$ denotes the video frame at position $(x,y)$ and time $t$. If the observed intensities change over time is denoted as a displacement function $\delta(t)$, such that $I(x,y,t) = f(x+\delta_x(t), y+\delta_y(t))$, where $\delta_x(t)$ and $\delta_y(t)$ are the displacement functions in $x$ and $y$ directions respectively. The goal of motion magnification can be expressed as follows:
\begin{equation}
\hat{I}(x,y,t) = f(x+(1+\alpha)\delta_x(t), y+(1+\alpha)\delta_y(t))
\end{equation}
where $\alpha$ is a magnification factor.
Under the first order Taylor series expansion about $x$ and $y$ directions,  the video $I$ can be rewritten as
\begin{equation}
{I}(x,y,t) \approx f(x,y)+ \delta_x(t)\frac{\partial{f(x,y)}}{\partial{x}} + \delta_y(t)\frac{\partial{f(x,y)}}{\partial{y}}
\end{equation}
Let $B(x,y,t)$ be the result of applying a broadband temporal bandpass filter to the input video $I$ at every position $(x,y)$, such that all the components except $f(x,y)$ are filtered. $B(x,y,t)$ can be represented as
\begin{equation}
B(x,y,t) = \delta_x(t)\frac{\partial{f(x,y)}}{\partial{x}} + \delta_y(t)\frac{\partial{f(x,y)}}{\partial{y}}
\end{equation}
Thus, the processed video $\hat{I}(x,y,t)$ can be expressed as
\begin{equation}
\hat{I}(x,y,t) = I(x,y,t) + \alpha B(x,y,t)
\end{equation}
Combining Eqs. 1,2,3 and 4, the motion magnified video $\hat{I}$ can be finally rewritten as
\begin{equation}
\hat{I}(x,y,t) = f(x,y) + (1+\alpha)[\delta_x(t)\frac{\partial{f(x,y)}}{\partial{x}} + \delta_y(t)\frac{\partial{f(x,y)}}{\partial{y}}]
\end{equation}
Compared Eq.2 and Eq.5, we could conclude that the spatial displacement $\delta(t)$ of the local image $f(x,t)$ at time $t$, has been amplified to a magnitude of $(1+\alpha)$. The effect of magnification is highly dependent on the filter and the magnification factor $\alpha$ used. In our work, the value of $\alpha$ is selected optimally by visual inspection of the proposed videos from the training dataset.
\begin{figure}[!t]
    \hspace{0.61em}
    \includegraphics[width =11.6cm, height=9cm]{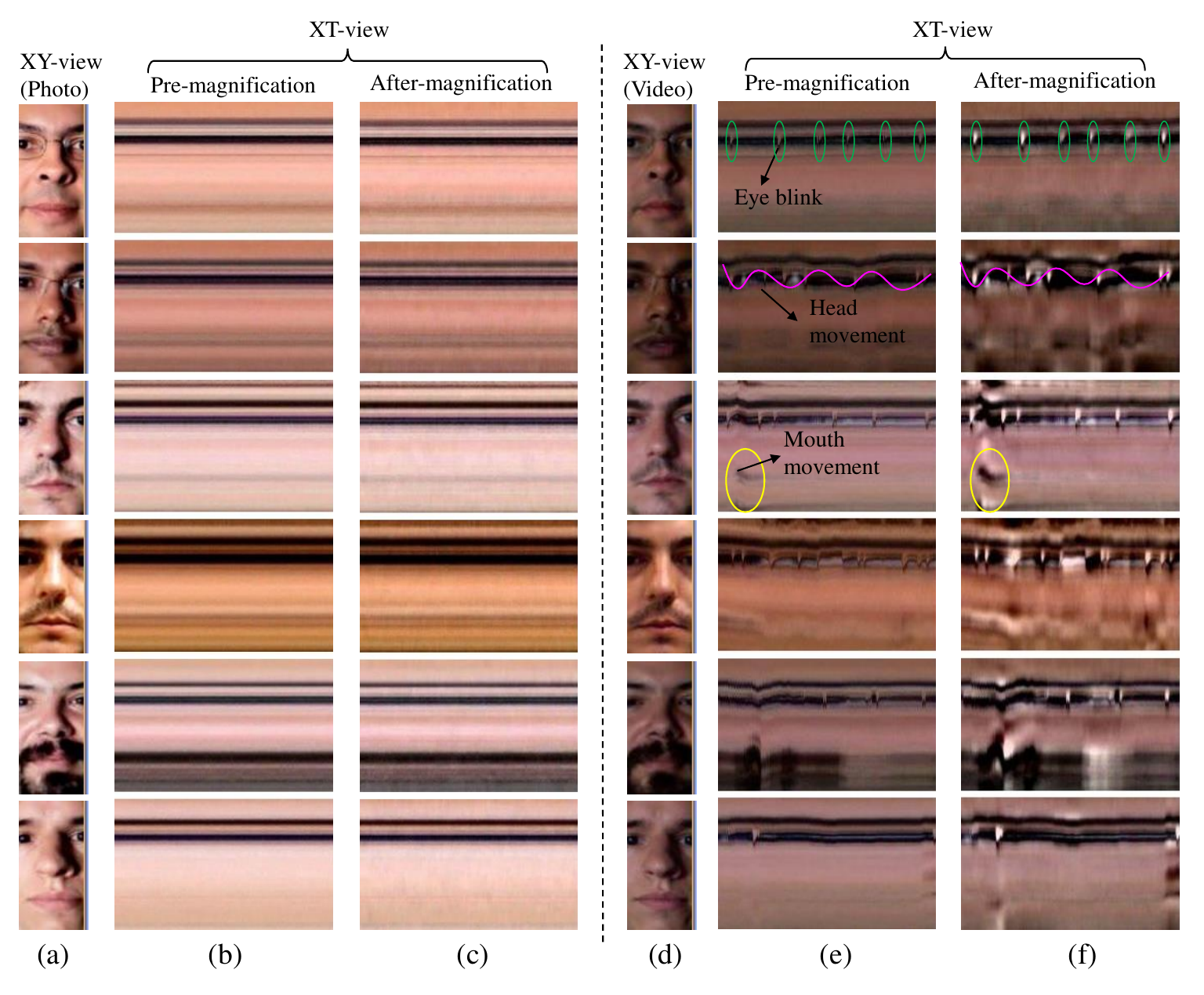}%

\begin{spacing}{0.9}
{\small{Fig.2: Column (a) shows the video of fixed photos in XY view, columns (b) and (c) are the corresponding images in XT view without and with magnification, respectively. Column (d) is the video with dynamic facial expressions in XY view, columns (e) and (f) represent the corresponding images in XT view without and with magnification, respectively.}}
\end{spacing}
\end{figure}

To demonstrate the results of this motion magnification preprocessing method, a video clip has been chosen and shown in Fig. 2 in XY-T and XT-Y view, where T represents the time axis. The original frames are taken at equal time intervals. As there are no motions contained in the static photo-video (Fig. 2(a)), no obvious changes in both the pre-magnification and after-magnification video can be observed in XT view (see Fig. 2(b) and (c)). However, for the video contains dynamic facial expressions (Fig. 2(d)), significant changes across frames can be observed in XT view (see Fig. 2(e) and (f)). Compared with the video pre-magnification, the changes on facial expression and head movement are much more significant in the video after-magnification (color marked regions in columns (e) and (f)).

\subsection{Spatial Feature Extraction}
To fully make use of static frame appearance, we used the VGG-16 \cite{Simonyan2014Very} based structure for spatial feature extraction. The VGG-16 is a very deep convolutional network with up to 16 weight layers (13 convolutional layers and 3 fully-connected layers). It takes $224 \times 224$ images as input, each convolution has a kernel of size $3 \times 3$, and all max-pooling layers are performed in a $2 \times 2$ window with stride 2. We used the pre-trained "VGG-Face" \cite{Parkhi2015Deep} model to initialize the parameters of the network. The "VGG-Face" was trained on a massive face data set of 2.6M images of 2622 subjects. We assume that both genuine and fake face images are involved in this data set, and use the idea of transfer learning to transfer the "knowledge" that learned by VGG model to our new task of face anti-spoofing. Transfer learning can be used to avoid overfitting in the training of our network, considering the current publicly available face spoofing datasets are too limited to train a generalized network.
\begin{figure*}[!h]
    \hspace{-1.5cm}
    \includegraphics[width = 14.5cm, height=7.3cm]{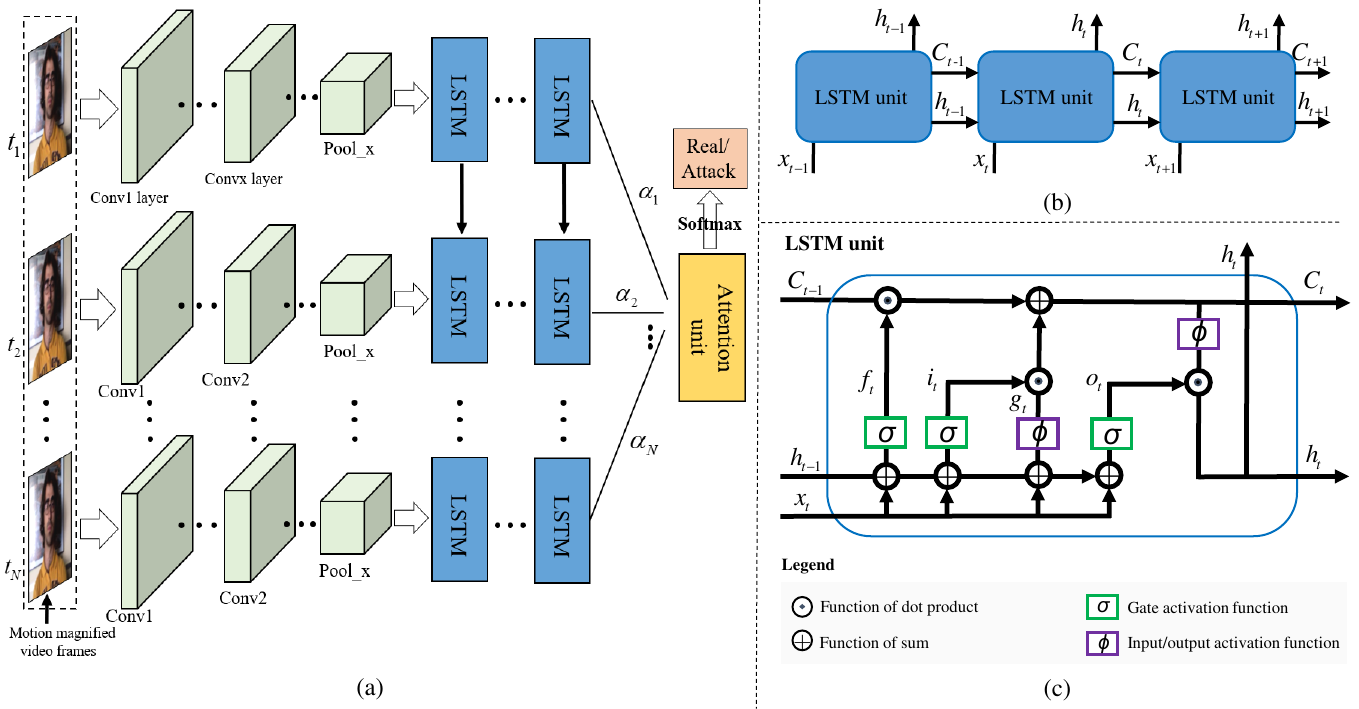}%

\begin{spacing}{0.9}
{\small{Fig.3: (a) The flowchart of the proposed CNN-LSTM framework. (b) The cascaded LSTM architecture. (c) Illustration of a single LSTM unit, the current state $t$ depends on the past state $t_1$ of the same neuron.}}
\end{spacing}
\end{figure*}

Similar with the work in \cite{Xu2016Learning}, we propose to have the CNNs underlying the LSTMs, therefore the local and dense property from convolution operation could be leveraged and the temporal feature across frames can be learned and stored in LSTM units. Fig. 3(a) shows the overall framework of the proposed joint CNN-LSTM model. The released ¡±VGG-Face¡± model was used to initialize the parameters of the network. Since the CNNs learn more generic features on the bottom of the network and more intricate, dataset-specific features near the top of the network \cite{Karpathy2014Large}, the pre-trained last 3 fully-connected layers were removed and only the convolutional and max-pooling layers were reserved. Another important reason we discarded the fully-connected layers is that the neuron activations from fully-connected layers do not encode spatial information, which is extremely important for capturing motion cues in the following LSTM architecture. Therefore, the last pooling layer of the CNN architecture is connected directly to the LSTM, the pooling layer contains rich spatial information which could be taken use of by the LSTM to explore the temporal features across continuous frames.

\subsection{The LSTM and Attention Mechanism}
Since the input of our architecture are videos which contain dynamic content, the variations between video frames may encode additional useful information for the differentiation of genuine and fake faces. To capture the temporal dynamic information across frames, we proposed to use the Long Short-Term Memory (LSTM) \cite{Gers2003Learning} architecture to explicitly consider sequences of CNN activations. The LSTM units can discover long-range temporal relationships from the input sequences by making use of the memory cells, which could store and output information.

As illustrated in Fig. 3(b) and Fig. 3(c), each LSTM unit has a memory cell ($C_t$) and three gates: the input gate ($i_t$), output gate ($o_t$) and forget gate ($f_t$). The memory cell ($C_t$) could store and output information, allowing it to better discover long-range temporal relationships. The gates serve to modulate the interactions between the memory cell itself and its environment. The input gate controls how much the input influence the internal state by multiplying the cell's non-linear transformation of inputs $g_t$. The output gate decides how much the internal state to transfer to the unit output. The forget gate can modulate the memory cell's self-recurrent connection, allowing the cell to remember or forget its previous state, as needed. The LSTM unit updates for timestep $t$ are:
\begin{equation}
f^t = \sigma(T_f x^t + R_f h^{t-1} + b_f)
\end{equation}
\begin{equation}
i^t = \sigma(T_i x^t + R_i h^{t-1} + b_i)
\end{equation}
\begin{equation}
g^t = \phi(T_g x^t + R_g h^{t-1} + b_g)
\end{equation}
\begin{equation}
C^t = g^t \odot i^t + C^{t-1} \odot f^t
\end{equation}
\begin{equation}
o^t = \sigma(T_ox^t + R_oh^{t-1} + b_o)
\end{equation}
\begin{equation}
h^t = \phi(C^t) \odot o^t
\end{equation}

For timestep $t$, $x^t$ and $h^t$ are the input and output, respectively. $T$ is the input weight matrix, $R$ is the recurrent weight matrix, and $b$ is the bias vector. $\sigma(x) = \frac{1}{1+e^-x}$ and $\phi(x)=\frac{e^x-e^{-x}}{e^x+e^{-x}}$ are the element-wise non-linear activation functions, mapping real values to (0, 1) and (-1,-1), respectively.

Along with the training of LSTM, we propose to encode the hidden sequences into a fixed-length vector $c = (x_1, x_2, ... x_N)$:
\begin{equation}
c = \sum_{i=1}^{N} \alpha_i h_i   \qquad  i  \in 1, ..., N
\end{equation}
where $h_i$ is the hidden state at time $t_i$, and the weight $\alpha_{i}$ is the corresponding weight mapping $h_i$ to vector $c$. At each time step , $\alpha_{i}$ is computed by
\begin{equation}
\alpha_{i} = \frac{\text{exp}(W_i^{\top} h_{i-1})}{\sum_{j=1}^N \text{exp}(W_j^{\top} h_{i-1})}
\end{equation}

where $\alpha_{i}$ can be thought of as the probability that reflecting the importance of the hidden state $h_i$. This attention mechanism decides which frames from the input sequences should be paid attention to.  After calculating these probabilities, the model could focus on the samples with dynamic changes instead of taking expectation over all the input sequences. The fixed-length vector $c$ is finally connected to the softmax layer to compute the class specific probabilities.

To balance the learning level of CNN and LSTM for the purpose of better generalization ability, we proposed a loss combining strategy to train the joint network, where the CNN batch loss and the LSTM loss was combined and trained simultaneously.

\section{Experiments}
In this section, we evaluate the performance of the proposed joint CNN-LSTM framework on two popular benchmark datasets. We first give a brief description of these two datasets. Performance evaluation of the proposed method along with several popular baseline methods are then reported using the same protocols.

\subsection{Experimental settings and Datasets}
All the input video frames are resized to $224 \times 224$. Instead of only using the face region as the input, we extend the border of the face region to include some background, as the relative movement between the face and the background is very important for LSTM to capture temporal information. The network is trained using the AdamOptimizer in tensorflow with the learning rate of 0.00001. Batch normalization layer is used to ensure inputs to a layer are normalized. The experiments were conducted on two up-to-date publicly available face anti-spoofing datasets, MSU-MFSD \cite{Wen2015Face} and Replay-Attack \cite{Chingovska2012On}. Following are the brief introductions of the two datasets:

\textsl{MSU-MFSD}: These dataset contains 280 video recordings of genuine and attack faces. 35 individuals have participated in the development of this database with a total of 280 videos. Two kinds of cameras with different resolutions ($720 \times 480$ and $640 \times 480$) were used to record the videos from the 35 individuals. For the real accesses, each individual has two video recordings captured with the Laptop cameras and Android, respectively. For the video attacks, two types of cameras, the iPhone and Canon  cameras were used to capture high definition videos on each of the subject. The videos taken with Canon camera were then replayed on iPad Air screen to generate the HD replay attacks while the videos recorded by the iPhone mobile were replayed itself to generate the mobile replay attacks. Photo attacks were produced by printing the 35 subjects' photos on A3 papers using HP colour printer. The recording videos with respect to the 35 individuals were divided into training (15 subjects with 120 videos) and testing (40 subjects with 160 videos) datasets, respectively. 30 real access videos and 90 attack videos are contained in the training dataset, while 40 real accesses and 120 attack videos are include in the testing dataset.

\textsl{Replay-Attack}: 50 subjects with a total of 1200 videos regarding 50 subjects are contained in this dataset. For each subject, 4 genuine video sequences are collected in front of controlled and adverse  backgrounds. The first condition was under uniform background and office lights turned on, while the second condition was under non-uniform background with the office lights off. As for attack videos, three spoofing devices are used, including print attack, video attack and digital photo attack. The spoofing sequences were captured from fixed support and hand-hold mediums under the adverse and controlled background, respectively. The evaluation protocol divides this dataset into training (360 videos), testing (480 videos) and development (360 videos) subdatasets. The training and development subdatasets each contains 60 real accesses videos and 300 attack videos, whereas the testing subdataset contains 80 real accesses and 400 attack videos.

\begin{table*}
{Table 1: Comparison results of intra-test on Repaly-attack and MSU-MFSD.}
\scalebox{0.7}[0.78]{
\hspace{-1.5cm}
\begin{tabular}{c|| c | c | c | c| c | c | c | c | c }
\Xhline{1.2pt}
\multicolumn{1}{c||}{\multirow{4}*{\large{Datasets}}}   & \multirow{4}*{\large{Metrics}} & \multicolumn{8}{c}{\multirow{2}*{\large{Methods}}} \\
\multicolumn{1}{c||}{}                         &                 & \multicolumn{2}{c}{}                       \\
\cline{3-10}
\multirow{2}*{}      & \multirow{2}*{} & \multirow{2}*{$LBP$} & \multirow{2}*{$LBP^M$} & \multirow{2}*{$LBP$-$TOP$} & \multirow{2}*{$LBP$-$TOP^M$} & \multirow{2}*{$CNN$} & \multirow{2}*{$CNN^M$} & \multirow{2}*{$CNN$-$LSTM$}& \multirow{2}*{$CNN$-$LSTM^M$} \\ [-0.0ex]
               &              &       &    &       &    &       &    &               &\\
\hline
Replay-attack  & HTER$_e$(\%) & 0.18 & 2.82&4.25&5.75&2.70&5.18&\textbf{0.00} &\textbf{0.00}\\
(controlled)   & ACC(\%)      & 99.80&97.05&95.00&92.86&96.11&92.59&\textbf{100} &\textbf{100}\\
\hline
Replay-attack  & HTER$_e$(\%) & 4.67 & 5.65&12.25&13.25&4.41&5.40&\ 4.49 & \textbf{3.53}\\ [0.3ex]
(adverse)      & ACC(\%)      & 95.06 & 94.02&90.0&87.86&94.47&93.70&\ 94.51 &\textbf{96.47}\\ [0.3ex]
\hline
MSU-           & HTER$_e$(\%) &11.63 &14.62&24.17&28.75&16.06&19.73& 13.74& \textbf{10.36}\\ [0.3ex]
MFSD           & ACC(\%)      &88.37 &85.38&85.0 &84.38&89.57&85.20&\ 86.26 &\textbf{89.64}\\ [0.3ex]
\Xhline{1.2pt}
\end{tabular}
}
\end{table*}

\begin{table*}
{Table 2: Comparison results of inter-test between different pairs of datasets. The training set of the first column are used to train the model, whereas the testing set of the second column are used for testing. The superscript $M$ means magnification operation, while superscript $A$ stands for attention mechanism.}
\scalebox{0.62}[0.68]{
\hspace{-1.5cm}
\begin{tabular}{c | c || c | c | c | c | c | c | c | c }
\Xhline{1.2pt}
\multicolumn{2}{c||}{\multirow{2}*{\large{Datasets}}}   & \multirow{4}*{\large{Metrics}} & \multicolumn{7}{c}{\multirow{2}*{\large{Methods}}} \\
\multicolumn{2}{c||}{}                         &                 & \multicolumn{2}{c}{}                       \\
\cline{1-2} \cline{4-10}
\multirow{2}*{Train}     & \multirow{2}*{Test} & \multirow{2}*{} & \multirow{2}*{$LBP$} & \multirow{2}*{$LBP$-$TOP$} & \multirow{2}*{$CNN$} & \multirow{2}*{$CNN$-$LSTM$} & \multirow{2}*{$CNN$-$LSTM^M$} & \multirow{2}*{$CNN$-$LSTM^A$} &  \multirow{2}*{$CNN$-$LSTM^{AM}$} \\ [-0.0ex]
               &               &              &       &    &       &    &       &    &               \\
\hline
Replay-attack  & Replay-attack & HTER$_e$(\%) &14.00  &\textbf{13.94}&25.38&31.32&23.10&23.70& 15.28 \\
(adverse)      &(controlled)   & ACC(\%)      &80.00  &80.98&73.91&78.63&80.66&80.60&\textbf{84.70} \\
\hline
Replay-attack  &   MSU-        & HTER$_e$(\%) &39.79  &32.93&48.60&32.43&30.15&29.52&\textbf{25.72} \\ [0.3ex]
(adverse)      &     MFSD      & ACC(\%)      &60.27  &64.38&50.16&68.98&70.99&71.15&\textbf{78.65} \\ [0.3ex]
\hline
MSU-           & Replay-attack & HTER$_e$(\%) &37.04  &30.50&37.11&23.63&14.75&18.97&\textbf{12.37} \\ [0.3ex]
MFSD           &(controlled)   & ACC(\%)      &62.96  &70.71&62.88&79.15&81.17&80.77&\textbf{82.32} \\ [0.3ex]
\Xhline{1.2pt}
\end{tabular}
}
\end{table*}

\subsection{Experimental Protocols and Performance Metric}
We followed the default protocols offered in this two databases, however, considering that no development set is contained in MSU-MFSD dataset, we equally split the testing set into a couple, the development set and the new testing set. To be specific, there are 30, 40 and 30 subjects used for training, testing and development in Replay-Attack database, respectively, while the training, testing and development datasets in MSU-MFSD each contains 30, 20 and 20 subjects, respectively.

To keep consistent to previous work, Half Total Error Rate (HTER) was used as the metric in our experiments. HTER is defined as:
\begin{equation}
HTER = \frac{FAR(\tau, D)+ FRR(\tau, D)}{2}
\end{equation}
where FAR is the measure of the likelihood that the biometric security system will incorrectly accept an access attempt by an unauthorized, and FRR is the measure of the likelihood that the biometric security system will incorrectly reject an access attempt by an authorized user. Since both False Acceptance Rate (FAR) and False Rejection Rate (FRR) depend on a certain threshold $\tau$, increasing the FAR will usually reduce the FRR and vice-versa. In this case, we used the development set to determine the threshold $\tau$ corresponding to Equal Error Rate (ERR) for the computing of HTER, following the previous works \cite{Yang2014Learn,Xu2016Learning}. The testing accuracy (ACC) achieved by each methods is also calculated for comparison.

Three popular methods, $LBP_{8, 1}^{u2}$ \cite{Maatta2011Face},  $LBP$-$TOP$\cite{Pereira2012LBP} and the standard CNN, are used to compare with the proposed approach on the same experimental protocols. The features captured by the $LBP_{8, 1}^{u2}$ and $LBP$-$TOP$ are all fed into SVM to obtain the final classification results. For the LBP method, all the consecutive frames of each video clip of the training set were used to extract the 59-dimensional holistic features, and to train the SVM classifier. The final results were achieved on the test set by averaging the probabilities of all the consecutive frames per video clip. The features of LBP-TOP were extracted per video clip, as this method involved the spatio-temporal information from video sequences. The traditional CNN is also implemented for the comparison with the proposed CNN-LSTM framework where the structure of CNN keep unchanged.

\subsection{Intra-test Restluts}
Table 1 illustrates the results of intra-test between different methods on the datasets of MSU-MFSD and Repaly-Atttack. As can be seen, all the methods listed achieve fairly good performance on the three datasets. For the methods LBP, LBP-LOP and standard CNN, the HTERs with magnification are relatively lower and the ACCs are higher than that without magnification. This is reasonable, as we found when applying the Eulerian algorithm for motion magnification, the image qualities of the magnified video clip were influenced although the motions across the frames were amplified. However, based on the comparisons between CNN-LSTM and CNN-LSTM$^{M}$, it's easy to draw the conclusion that the motion magnification significantly improves the performance of CNN-LSTM joint framework, this is mainly because more motion information were perceived and leveraged by the LSTM when applying motion magnification on the input video clips. It is also important to take note of the results of CNN on Replay-attack (controlled) and Replay-attack (adverse), where we observe a relatively worse performance even compared with artificial features LBP and LBP-TOP. Actually, in the training phase, the CNN accuracy on this two datasets was nearly 100 percent. The degraded performance is mainly attributed to over-fitting, considering the images contained in Replay-attack (controlled) and Replay-attack (adverse) are too limited to tain a generalized deep network. However, the training and testing images used in CNN-LSTM is the same with the method of CNN, while this over-fitting problem doesn't exist, indicating that the joint CNN-LSTM framework can effectively overcome the overfitting problem, since more cues can be used for the discrimination of genuine and fake face images.

\subsection{Inter-test Results}
To evaluate the generalization ability of the proposed method, we conduct inter-test on the three datasets. Dataset A is firstly used to train the model and dataset B is used for testing. The inter-test results are reported in table 2. As can be observed, among all the methods listed in Talbe 2, CNN-LSTM with both magnification and attention mechanism achieve the best inter-test performance in general. The worst inter-test results are achieve by LBP and CNN, since these two methods only keep focus on static picture information, non of the features regarding motions are used. If external environment (such as background, lighting, camera resolution, etc.) is changed, these static features may be totally different. For approaches LBP-TOP and CNN-LSTM, as these methods exploit more generalized features: (temporal information across frames) which does not alter with the change of external environment, the inter-test results reported by these approaches are much better than LBP and CNN. The performance achieved by CNN-LSTM$^M$ and CNN-LSTM$^A$ are both slightly better than CNN-LSTM, suggesting that the video magnification preprocess and attention mechanism are both contribute to the distinction of genuine/fake faces. With the video magnification some subtle motions would be amplified to be perceived by LSTM, and with the attention mechanism the LSTM can concentrate mainly on key frames that contain dynamic information.

For Replay-Attack-adverse and Replay-Attack-controlled datasets, only the illumination condition is different, other factors like spoofing materials and individuals are keep unchanged, therefore the domain shift is not serious between these two datasets. This is the reason why all the approaches listed achieved the best performance by testing across these two datasets. However, the MSU-MSFD database is totally different from Replay-Attack database, where the spoofing materials, individuals and illumination conditions are all different. As a result, the differences between these two databases are quite serious, and the performance degenerate significantly when inter testing cross these two datasets.

\section{Conclusions}
Due to the variety of materials among spoofing devices, current face anti-spoofing approaches usually suffer from poor generalization. The generalization ability of face anti-spoofing needs to be significantly improved before they can be adopted by practical application systems. In this paper, we keep focus on a more generalized features, the fine-grained motions across video frames, for the purposed of robust face anti-spoofing. The LSTM is used to extract temporal features from the input videos, while eulerian motion magnification and attention mechanism are proposed to ensure LSTM make full of the dynamic changes exhibited by individuals. To demonstrate the effectiveness of the proposed method, intra-test and inter-test are performed on two challenging datasets, and several popular methods are implemented for comparison. The experimental results verify that this dynamic changes contained in the temporal features are quite helpful to improve the generalization ability of the proposed approach.

\section{Conflict of interest}
None.

\section{Acknowledgements}
This work was supported in part by the 111 project (B14039), the Fundamental Research Funds for the Central Universities of China (No. A03017023701112) and the funding from the Center for Information in Medicine in UESTC (No. ZYGX2015J138). The content is solely the responsibility of the authors

\bibliography{mybibfile}

\end{document}